\documentclass{article}

\usepackage[preprint,nonatbib]{neurips_2024}

\usepackage[utf8]{inputenc} 
\usepackage[T1]{fontenc}    
\usepackage{hyperref}       
\usepackage{url}            
\usepackage{booktabs}       
\usepackage{amsfonts}       
\usepackage{nicefrac}       
\usepackage{microtype}      
\usepackage{xcolor}         
\usepackage{amsmath}
\usepackage{graphicx}
\usepackage{multirow}
\usepackage{array}
\usepackage{tabularx} 
\usepackage{enumitem}
\usepackage{wrapfig}
\usepackage{caption}
\usepackage{lipsum}
\usepackage{algorithm}
\usepackage{algpseudocode}
\usepackage{xcolor} 
\usepackage{cleveref}
\title{Uncertainty-guided Optimal Transport in \\ Depth Supervised Sparse-View 3D Gaussian}
\author{%
  Wei Sun \\
  UCAS\\
  \texttt{sunwei162@mails.ucas.ac.cn} \\
  \And
  Qi Zhang \\
  UCAS \\
    \texttt{zhangqi203@mails.ucas.ac.cn} \\
  \And
  Yanzhao Zhou \\
  UCAS\\
  \texttt{zhouyanzhao@ucas.ac.cn} \\
  \And 
  Qixiang Ye \\
  UCAS \\
  \texttt{qxye@ucas.ac.cn} \\
  \And
  Jianbin Jiao \\
  UCAS \\
  \texttt{jiaojb@ucas.ac.cn} \\
  \And
  Yuan Li\thanks{Corresponding author.} \\
  UCAS \\
  \texttt{liyuan23@ucas.ac.cn} \\
}

\begin{document}

\maketitle
\begin{abstract}
3D Gaussian splatting has demonstrated impressive performance in real-time novel view synthesis. However, achieving successful reconstruction from RGB images generally requires multiple input views captured under static conditions. To address the challenge of sparse input views, previous approaches have incorporated depth supervision into the training of 3D Gaussians to mitigate overfitting, using dense predictions from pretrained depth networks as pseudo-ground truth. Nevertheless, depth predictions from monocular depth estimation models inherently exhibit significant uncertainty in specific areas. Relying solely on pixel-wise L2 loss may inadvertently incorporate detrimental noise from these uncertain areas. In this work, we introduce a novel method to supervise the depth distribution of 3D Gaussians, utilizing depth priors with integrated uncertainty estimates. To address these localized errors in depth predictions, we integrate a patch-wise optimal transport strategy to complement traditional L2 loss in depth supervision. Extensive experiments conducted on the LLFF, DTU, and Blender datasets demonstrate that our approach, UGOT, achieves superior novel view synthesis and consistently outperforms state-of-the-art methods.
\end{abstract}

\section{Introduction}

Novel View Synthesis (NVS) has emerged as a crucial task in 3D computer vision, underpinning advancements in applications ranging from virtual reality to image editing. NVS aims to generate imagery from any viewpoint within a scene, which typically requires meticulous modeling based on multiple scene images. Leveraging implicit scene representations and differentiable volume rendering, Neural Radiance Fields (NeRF)~\cite{mildenhall2021nerf} and its derivatives have shown significant progress in this area. However, the NeRF framework is hampered by extensive training and rendering times. While various NeRF variants have managed to accelerate these processes, they often compromise the image quality, particularly in high-resolution renderings.

As an effective alternative, 3D Gaussian splatting (3D-GS)~\cite{kerbl20233d} has gained attention for its exceptional training and inference speeds while preserving quality competitive with NeRF. This method employs anisotropic 3D Gaussians and adaptive density controls, enabling precise and explicit scene representations ideal for NVS. Unlike NeRF’s cumbersome volume rendering, 3D-GS utilizes efficient splatting that projects 3D Gaussians onto a 2D plane, facilitating real-time rendering. However, this technique can lead to over-reconstruction in scenes with sparse views due to its localized approach, where large Gaussians might dominate rendering, causing blur and artifacts as well as frequency discrepancies compared to the ground truth.

Previous methods~\cite{chung2023depth, li2024dngaussian} have utilized pixel-wise L2 loss to supervise depth in Gaussian-rendered outputs without considering the variability in depth estimation accuracy across different areas, which can result in misinterpreted geometries. NeRF models, which use implicit representations, benefit from the inherent smoothness of MLPs, thus minor discrepancies in depth do not significantly impact the final rendered image~\cite{rau2024depth}. However, explicit point-based methods like 3D-GS can exacerbate these inaccuracies. Given these challenges, our analysis leads to four key insights:
\begin{enumerate}
    \item The depth of each pixel in the final render is often determined by a subset of Gaussian splats with the highest weights, negating the need to optimize all Gaussians based on ground truth depth, which can destabilize training due to depth error.
    \item The depth estimation's uncertainty should be explicitly modeled at each image location, thereby primarily utilizing the more reliable depth estimates from areas with lower uncertainty.
    \item The L2 loss enforces the rendered depth to exactly replicate the estimated depth prior, potentially leading to overfitting. By supervising the depth distribution of each Gaussian, it is possible to mitigate the impact of noisy depth estimation on scene geometry.
    \item Pixel-wise loss tends to exacerbate geometric degradation in areas with high depth uncertainty; therefore, a regional supervision approach should also be adopted to minimize this effect.
\end{enumerate}

Building on these insights, we propose an Uncertainty-guided Optimal Transport (UGOT) approach to optimize the depth distribution of 3DGS patch-wisely. Utilizing off-the-shelf pre-trained generative diffusion models as depth priors, we enhance depth supervision with uncertainty estimates derived during the denoising process. This method prioritizes the use of more reliable depth estimates from areas with lower uncertainty. Additionally, we employ differentiable sampling of the depth distribution from the most influential Gaussian splats and their participation weights. In each iteration, we define random-sized patches, compute patch mean for each set of samples, and incorporate an optimal transport strategy to construct a regional depth distribution loss.

\begin{figure}[t]
    \centering
    \includegraphics[width=\linewidth]{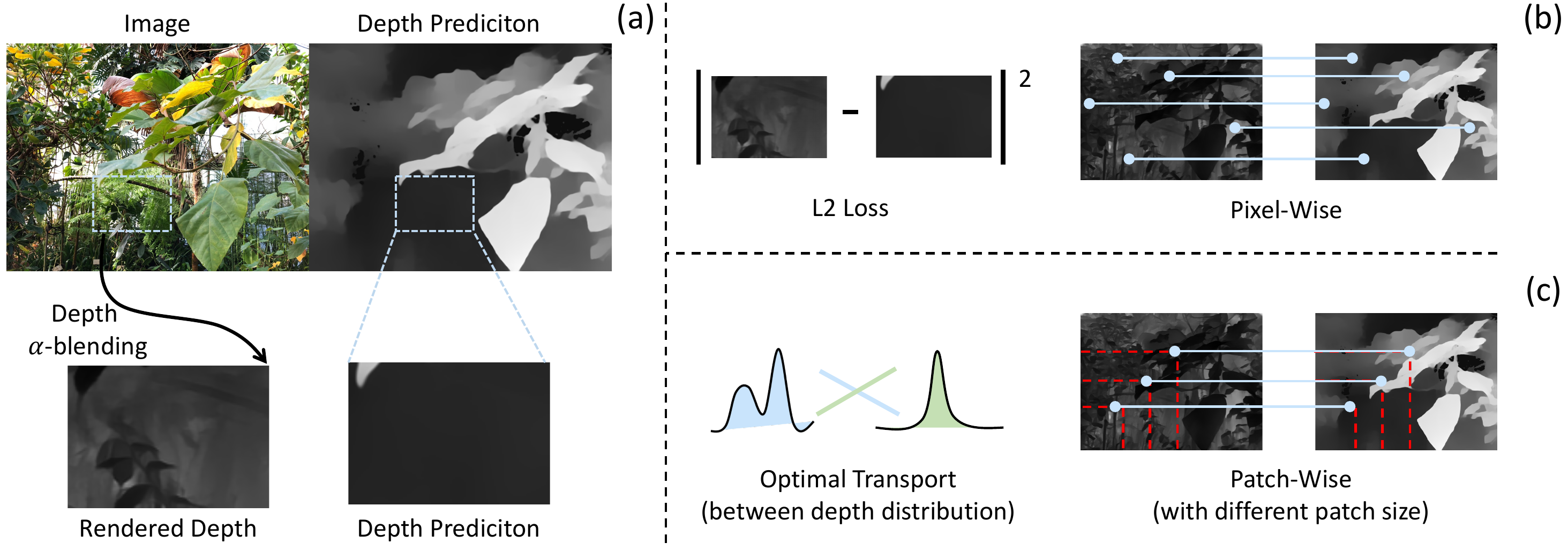}
    \caption{The comparison with the previous method.}
    \label{fig:teaser}
\end{figure}

\section{Related Work}

\textbf{Novel-view Synthesis}.
Structure from Motion (SfM)~\cite{ullman1979interpretation} and Multi-View Stereo (MVS)~\cite{tomasi1992shape} are traditional techniques used to reconstruct 3D structures from multiple images and have long been a focus in the computer vision field. For achieving denser and more realistic reconstructions, deep learning-based 3D reconstruction methods have gained prominence~\cite{han2019image, mildenhall2021nerf, xie2022neural}. Among these, Neural Radiance Fields (NeRF)~\cite{mildenhall2021nerf} stands out as a notable method, utilizing a neural network to represent 3D scenes. NeRF employs an MLP network for 3D space representation and volume rendering, leading to numerous follow-up studies in 3D reconstruction~\cite{barron2021mip, barron2022mip, gao2022nerf, tewari2022advances, wang2021neus, yariv2021volume}. To address the slow rendering speed of NeRF, considerable efforts have been made to achieve real-time rendering through explicit representations such as sparse voxels~\cite{fridovich2022plenoxels, liu2020neural, sun2022direct, yu2021plenoctrees}, featured point clouds~\cite{xu2022point}, tensors~\cite{chen2022tensorf}, and polygons~\cite{chen2023mobilenerf}. These representations are composed of local elements that function independently, enabling fast rendering and optimization. Building on this idea, various other representations have been explored, including Multi-Level Hierarchies~\cite{muller2021real, muller2022instant}, infinitesimal networks~\cite{garbin2021fastnerf, reiser2021kilonerf}, and triplanes~\cite{chan2022efficient}. Among these approaches, 3D Gaussian Splatting~\cite{kerbl20233d} stands out by representing radiance fields with a set of anisotropic 3D Gaussians and rendering them using differentiable splatting. This technique has achieved significant success in fast and high-quality reconstruction of complex real scenes. While 3D Gaussian Splatting performs excellently with dense input views and has shown success in various 3D tasks~\cite{liu2023zero, tang2023dreamgaussian, wu20234d}, its effectiveness in scenarios with sparse view inputs remains an open challenge.

\textbf{Few-Shot Novel-view Synthesis}.
The objective of few-shot novel view synthesis is to generate new views from a limited number of sparse input views. Various approaches have been explored to tackle this challenge, with some focusing on regularization strategies specifically tailored for NeRF implementations~\cite{deng2022depth, kim2022infonerf, niemeyer2022regnerf, yang2023freenerf}. Others have ventured into using pre-trained generative models for data augmentation~\cite{chen2021mvsnerf, cong2023enhancing, kulhanek2022viewformer, yu2021pixelnerf, zhou2023sparsefusion}, while additional strategies involve leveraging pre-trained models to provide data-driven priors that guide the training regimen~\cite{jain2021putting, wynn2023diffusionerf}. Another significant approach is depth distillation~\cite{deng2022depth, roessle2022dense, song2024darf, wang2023sparsenerf}, which proves effective in sparse-view neural field applications. Nonetheless, many techniques falter in the context of 3D Gaussian Splatting (3DGS) due to its pronounced locality and the absence of MLP smoothing capabilities. This often results in the existing per-pixel regularization methods failing to effectively address floating artifacts observed when only a few images are available~\cite{chung2023depth}.

\textbf{Depth Supervision in Novel-view Synthesis}.
    Depth cues, long-established as pivotal in various 3D vision applications~\cite{wang2022uncertainty, wang2021multi, wang2024contrastive, wang2024robust, zhang2022revisiting}, leading another line of work in guiding novel-view synthesis. Two primary approaches emerge in leveraging depth information. The first approach~\cite{deng2022depth, roessle2022dense} extracts precise yet sparse depth values from trustworthy point clouds, while the second approach~\cite{hu2023consistentnerf, song2024darf, uy2023scade, wang2023sparsenerf, yu2022monosdf} derives depth insights from advanced monocular depth estimators\cite{ranftl2021vision, ranftl2020towards, uy2023scade}. Although monocular depth estimation addresses the scarcity of point clouds in sparse-view settings, its out-of-domain application introduces challenges such as partial occlusions, shading, and reflections, often resulting in distorted geometries.To mitigate the inherent inaccuracies of monocular depths, prior works in sparse-view synthesis have adopted various scale-invariant losses~\cite{deng2023nerdi, song2024darf, yu2022monosdf, zhu2023fsgs}, including depth ranking losses~\cite{wang2023sparsenerf, xu2023neurallift}. Furthermore, efforts have been made to model noise and manage uncertain regions~\cite{deng2022depth, roessle2022dense}, alongside the introduction of softer constraints~\cite{wang2023sparsenerf} aimed at reducing noise impact. However, these methods do not adequately address the issue of noise within depth-supervised 3D Gaussian contexts. Specifically, the adaptable nature of Gaussians to erroneous depth cues necessitates additional regularization strategies. Moreover, aligning depth to a uniform scale may disregard the influence of noise across various scales. Such oversight can result in a noisy distribution of primitives, especially noticeable in areas with complex textures.

\section{Method}

\begin{figure}[t]
    \centering
    \includegraphics[width=\linewidth]{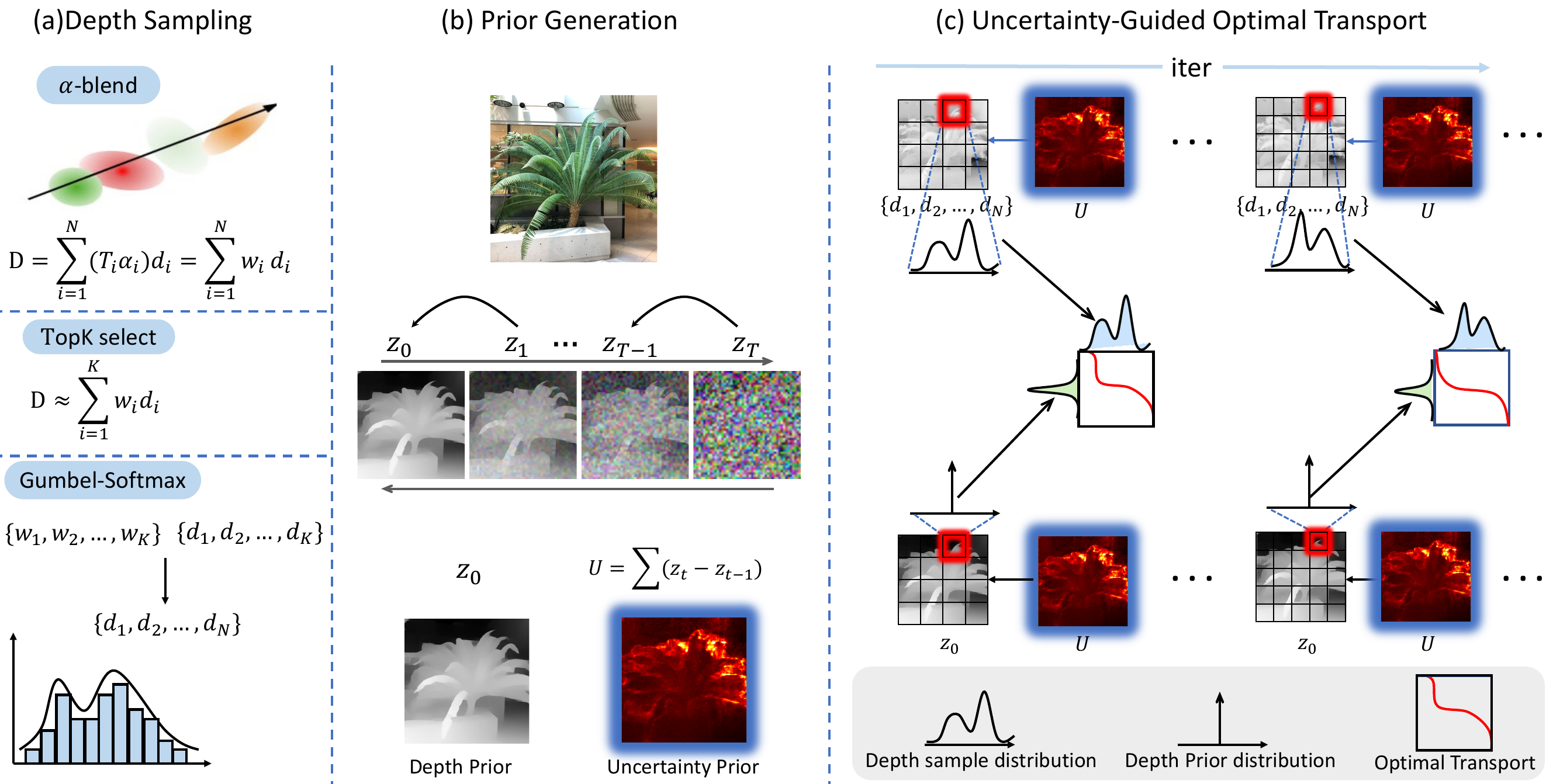}
    \caption{Overview of our method design.}
    \label{fig:overview}
\end{figure}

An overview of our method is shown in ~\Cref{fig:overview}.

\subsection{Preliminaries of 3D Gaussian Splatting} \label{subsec:preliminary}
3D Gaussian Splatting (3DGS)~\cite{kerbl20233d} represents a 3D scene using a suite of 3D Gaussians. The properties of each Gaussian include a 3D position $\mathbf{p} = \{x, y, z \} \in \mathbb{R}^3$, a 3D size scaling factor $\mathbf{s} \in \mathbb{R}^3$, a rotation quaternion $\mathbf{q} \in \mathbb{R}^4$, a color $\mathbf{c} \in \mathbb{R}^3$, and an opacity value $o_i \in \mathbb{R}$. These parameters are learnable and can be collectively symbolized by $\Gamma_{{\theta}_{i}} = \{\mathbf{p}_{i}, \mathbf{s}_{i}, \mathbf{q}_{i}, \mathbf{c}_{i}, o_{i} \}$, where $i$ denotes the $i$-th Gaussian. 
Specifically, for computing the pixel color ${C}$, it utilizes $\alpha$-blending point-based rendering by blending $\mathcal{N}$ points in the front-to-back depth order: 
\begin{equation}
{C} = \sum_{i \in \mathcal{N}} T_i \alpha_i \mathbf{c}_i, 
\end{equation}
where $\mathcal{N}$ is the set of Gaussian points that overlap with the given pixel. 
$\alpha_{i}$ is calculated by $\alpha_{i} = o_i G_i^{2D}$, where $G_i^{2D}$ denotes the function of the $i$-th Gaussian projected onto 2D plane. 
The transmittance $T_i$ is calculated as the product of opacity values of preceding Gaussians that overlap the same pixel: $T_i = \prod_{j=1}^{i-1} (1 - \alpha_j)$. 
The initial 3D Gaussians are constructed from the sparse data points created by Structure-from-Motion (SfM)~\cite{ullman1979interpretation} using COLMAP~\cite{schonberger2016structure}. 
To optimize these 3D Gaussians, Gaussian Splatting employs a differentiable rendering technique for projecting them onto the 2D image plane and utilizes gradient-based color supervision to update parameters.  
The reconstruction loss is computed by minimizing the rendered image $\hat{I}$ and the ground truth image $I$ color, which is formulated as: 
\begin{equation}
\mathcal{L}_{rgb} = (1-\lambda)\mathcal{L}_{1}(\hat{I}, I) + \lambda\mathcal{L}_{D-SSIM}(\hat{I}, I), 
\end{equation}
where $\lambda$ is set to 0.2. 
3DGS has proven effective in 3D reconstruction tasks, showing more efficient inference speeds with high-quality reconstruction comparable to NeRF.

\subsection{Optimal Transport (OT)} measures the minimal cost to transport between two probability distributions~\cite{peyre2019computational, benamou2015iterative, chizat2018scaling, courty2016optimal, genevay2016stochastic}. We only provide a brief introduction to OT for discrete distributions and refer the readers to~\cite{peyre2019computational} for more details. Denote two discrete probability distributions \( p = \sum_{i=1}^n a_i \delta_{x_i} \) and \( q = \sum_{j=1}^m b_j \delta_{y_j} \), where both \( a \) and \( b \) are discrete probability vectors summing to 1, \( x_i \) and \( y_j \) are the supports of the two distributions respectively, and \( \delta \) is a Dirac function. Then the OT distance is formulated as follows: \( OT(p, q) = \min_{T \in \Pi(p,q)} \langle T, C \rangle \), where \( C \in \mathbb{R}^{n \times m}_{\geq 0} \) is the cost matrix with element \( C_{ij} = C(x_i, y_j) \) which reflects the cost between \( x_i \) and \( y_j \) and the transport probability matrix \( T \in \mathbb{R}^{n \times m}_{\geq 0} \) is subject to \( \Pi(p, q) = \{ T \mid \sum_{i=1}^n T_{ij} = b_j, \sum_{j=1}^m T_{ij} = a_i \} \). The optimization problem above is often adapted to include a popular entropic regularization term \( H = -\sum_{ij} T_{ij} \ln T_{ij} \) for reducing the computational cost, denoted as Sinkhorn algorithm~\cite{cuturi2013sinkhorn}.

\subsubsection{Depth Sampling}

In the point-based Gaussian rendering process, each pixel corresponds to a large number of Gaussian primitives that need optimization. However, a significant portion of the information is concentrated in the Gaussians with the highest weights. By extracting the top \( K \) Gaussians depth $\tilde{D}_{\text{top}} = \{d_1, d_2, \ldots, d_K\}$ for each pixel based on their weights $ \tilde{W}_{\text{top}} = \{w_1, w_2, \ldots, w_K\}$, we can reduce the computational and storage burden in distribution optimization, mitigate the drift of smaller Gaussians, improve optimization stability, and enhance overall rendering quality. Specifically, we extract the top \( K \) Gaussians for each pixel during the forward pass and sample discrete depth distributions based on their weights and depths using Gumbel-Softmax sampling:

\begin{equation}
p_j = \frac{\exp((\log(w_j) + g_j) / \tau)}{\sum_{k=1}^{K} \exp((\log(w_k) + g_k) / \tau)}
\label{eq:gumbel}
\end{equation}
\begin{equation}
d^{*}_1, d^{*}_2, \ldots, d^{*}_n = \sum_{j=1}^{K} p_j d_j, \quad \text{for } i = 1, 2, \ldots, n.
\end{equation}

where \( p_j \) is the softmax probability for each depth value \( d_j \), calculated using Gumbel noise and temperature \( \tau \), \( g_j \) are independent and identically distributed (i.i.d) Gumbel noise variables, \( \tau \) is the temperature parameter controlling the softness of the softmax distribution, and \( d^{*}_1, d^{*}_2, \ldots, d^{*}_n \) are the sampled depth values for each pixel, where \( n \) is the number of samples.

\subsubsection{Optimal Transport for Depth Sampling}

In our method, we utilize optimal transport (OT) to compare the sampled depth distribution \( D^* \) with the ground truth depth \( D_{\text{gt}} \) (single value, thus Dirac distribution) provided by a monocular depth estimation model. Given the discrete nature of the sampled depths and the Dirac delta distribution of the ground truth, we formulate the optimal transport problem as follows:

\begin{equation}
OT_\epsilon (D^*, D_{\text{gt}}) \stackrel{\text{def}}{=} \min_{T_{i} \in \Pi(D^*, D_{\text{gt}})} \sum_{i} C_{i} T_{i} - \epsilon \left[ \sum_{i} T_{i} \ln T_{i} \right]
\end{equation}

\begin{equation}
\Pi(D^*, D_{\text{gt}}) := \left\{ T_{i} \ \middle| \ \sum_{i} T_{i} = 1, \ T_{i} = w_i \right\}
\end{equation}

where \( D^* = \{d_1^*, d_2^*, \ldots, d_n^*\} \) is the set of sampled depth values, \( D_{\text{gt}} \) is the ground truth depth value from the monocular depth estimation model, \( \epsilon \) is the hyper-parameter for the entropic constraint, \( C_{i} \) is the cost function measuring the difference between \( d_i^* \) and \( D_{\text{gt}} \), \( T_{i} \) is the transport probability matrix, and \( \Pi(D^*, D_{\text{gt}}) \) is the set of all valid transport plans \( T_{i} \) that satisfy the marginal constraints.

\subsubsection{Patch-wise Optimal Transport}

To further enhance the robustness of depth optimization, we introduce a method that aggregates depth estimates over patches. During each iteration, a random patch size from \( s_1 \) to \( s_2 \) is selected. Let \( \bar{D}_{\text{patch}} = \{\bar{d}_1, \bar{d}_2, \ldots, \bar{d}_n\} \) represent the set of mean depth values for a single patch, where each \( \bar{d}_i \) is the mean of the \( i\)-th sampled depths within this patch. The optimal transport (OT) problem for each patch is then formulated as:

\begin{equation}
\text{OT}_{\epsilon}(\bar{D}_{\text{patch}}, \bar{D}_{gt}) = \min_{T \in \Pi(\bar{D}_{\text{patch}}, \bar{D}_{gt})} \sum_{i} C_{i} T_{i} - \epsilon \sum_{i} T_{i} \ln T_{i}
\end{equation}
\begin{equation}
\Pi(\bar{D}_{\text{patch}}, \bar{D}_{gt}) := \left\{ T_{i} \mid \sum_{i} T_{i} = 1, \ T_{i} = \bar{w}_i \right\}
\end{equation}
Here, \( \Pi(\bar{D}_{\text{patch}}, \bar{D}_{gt}) \) is the set of admissible transport plans between the depth sample means and their corresponding ground truth mean, \( \bar{w}_i \) represents the    patch mean weight of the \( i \)-th group of samples and \( \bar{D}_{gt} \) represents the patch mean ground truth depth.

\subsubsection{Uncertainty-Guided Optimal Transport}

As monocular depth estimation is inherently uncertain, it is crucial to measure the confidence in these predictions. While many methods exhibit promising depth prediction capabilities, generative models, especially denoising diffusion models, explicitly illustrates the role of uncertainty in the depth generation process. We adopt the DiffDP model~\cite{ji2023ddp}, a state-of-the-art diffusion model for depth estimation, which uses image features to denoise noisy depth maps. During inference, features are extracted from the input image, concatenated with random noise, and fed into a lightweight decoder to produce a final depth prediction. Given an input image \( \mathbf{I} \), DiffDP formulates the denoising process as:

\begin{equation}
p_\theta(z_0 \mid \mathbf{I}) = p_\theta(z_T) \prod_{t=1}^{T} p_\theta(z_{t-1} \mid z_t, \mathbf{I})
\end{equation}

where \( z_t \sim \mathcal{N}(0, \mathbf{I}) \) and \( z_0 \) corresponds to the final depth estimate from the model.

During the denoising process, the model updates its estimate recursively. Pixels with higher uncertainty will be updated more frequently, reflecting their instability. This iterative update serves as a proxy for measuring uncertainty in depth estimation. Instead of focusing solely on areas of high error, it captures the intrinsic uncertainty in the depth generation process.

To quantify the uncertainty of a depth estimate \( u(z_0) \), we compare each estimate to the previous one at each time step \( t \) and compute the average count of significant changes:

\begin{equation}
c(z_0:T) = \frac{1}{T} \sum_{t=T,\ldots,1} \left|z_t - z_{t-1}\right| 
\end{equation}

where \( z_t \) is the depth prediction at the \( T-t \) step.

Let \( \mathcal{M}(\cdot) \) denote a function that mirrors an image. The uncertainty for an image \( \mathbf{I} \) is then given by:

\begin{equation}
U(z_0:T \mid \mathbf{I}) = \frac{c(z_0:T \mid \mathbf{I}) + \mathcal{M}(c(z_0:T \mid \mathcal{M}(\mathbf{I}))) }{2}
\end{equation}

During each iteration, a random patch size \( w \times w \) is selected for analysis. The weight maps are normalized within each patch using softmax:

\begin{equation}
W_{\text{patch}}(i) = \text{Softmax}(-U_{\text{patch}}(i))
\end{equation}

The weighted mean depth values for the sampled depths and ground truth depths are then computed for each patch:

\begin{equation}
\overline{D}_{\text{patch}}^{U} = \overline{D}_{\text{patch}} \odot W_{\text{patch}}
\end{equation}
\begin{equation}
\overline{D}_{\text{gt}}^{U} = \overline{D}_{\text{gt}} \odot W_{\text{patch}}
\end{equation}

Finally, the optimal transport problem for the uncertainty-guided mean depths is formulated as:

\begin{equation}
OT_\epsilon (\overline{D}_{\text{patch}}^{U}, \overline{D}_{\text{gt}}^{U}) = \min_{T \in \Pi(\overline{D}_{\text{patch}}^{U}, \overline{D}_{\text{gt}}^{U})} \sum_i C_i T_i - \epsilon \sum_i T_i \ln T_i
\end{equation}

This framework integrates uncertainty into the depth optimization process, providing more robust depth supervision through meticulously designed weighting, particularly by reducing the impact of depth estimation in uncertain areas.

\subsubsection{Training Details}
In the supervision of depth estimation, we adopt the per-pixel $L_2$ loss on normalized patches from DNGaussian~\cite{li2024dngaussian}, referred to as $L_{\text{dn}}$. Additionally, we employ our proposed uncertainty-guided optimal transport (UGOT), optimizing the cost between distributions using the optimal transport distance, denoted as $L_{\text{ot}}$. 

The overall loss function combines these individual components with respective weights:

\begin{equation}
L = \lambda_1 \cdot L_{\text{dn}} + \lambda_2 \cdot L_{\text{ot}} + \lambda_3 \cdot L_{\text{color}} 
\end{equation}

Where $L_{\text{color}}$ is the RGB-losses in 3D gaussian~\cite{kerbl20233d}. By integrating optimal transport loss with per-pixel loss, we utilize the geometric priors within the patch context, while avoiding strictly enforcing rendered depth to exactly replicate depth prediction.

\section{Experiments}

\subsection{Setups}

\paragraph{Datasets} We conduct our experiment on three datasets: the NeRF Blender Synthetic dataset (Blender)~\cite{mildenhall2021nerf}, the DTU dataset~\cite{jensen2014large}, and the LLFF dataset~\cite{mildenhall2019local}. We adhere to the split settings used in previous works~\cite{niemeyer2022regnerf, wang2023sparsenerf, yang2023freenerf, li2024dngaussian} to train the model on 3 views and test on another set of images. To minimize background noise and focus on the target object, we apply object masks as used previously~\cite{niemeyer2022regnerf} for DTU during evaluations. For Blender, our approach is aligned with DietNeRF~\cite{jain2021putting} and FreeNeRF~\cite{yang2023freenerf}, training on 8 views and testing on 25 unseen images. Downsampling rates of 8, 4, and 2 are utilized for LLFF, DTU, and Blender respectively. Camera poses are assumed to be known via calibration or other methods.

\paragraph{Evaluation Metrics} Our evaluation framework utilizes PSNR, SSIM~\cite{wang2004image}, and LPIPS~\cite{zhang2018unreasonable} scores to quantitatively assess the reconstruction performance. 

\paragraph{Baselines} We compare our results with various current state-of-the-art (SOTA) methods including SRF~\cite{chibane2021stereo}, PixelNeRF~\cite{yu2021pixelnerf}, MVSNerf~\cite{chen2021mvsnerf}, Mip-NeRF~\cite{barron2021mip}, DietNeRF~\cite{jain2021putting}, RegNeRF~\cite{niemeyer2022regnerf}, FreeNeRF~\cite{yang2023freenerf}, and SparseNeRF~\cite{wang2023sparsenerf}. Direct comparisons are made with the best quantitative outcomes reported in the literature for these NeRF-based methods. The performance of 3D Gaussian Splattering (3DGS)~\cite{kerbl20233d} is also included.

\paragraph{Implementation Details} Our models are constructed using the official PyTorch 3D Gaussian Splattering codebase. We train the models for 6,000 iterations across all datasets. Parameters $\gamma = 0.1$ and $\tau = 0.95$ are set in the loss functions for all experiments. The hash encoder~\cite{li2024dngaussian} based neural renderer from~\cite{li2024dngaussian} is adopted in all 3D Gaussian model for fair comparison. DiffDP~\cite{ji2023ddp} is employed to predict monocular depth maps for all input views. The models of 3DGS and DNGaussian are initialized with a uniform distribution.

\begin{table}[t]\small
\centering
\caption{Rendering results. {\bf Bold} and \underline{Underline} indicate state-of-the-art (SOTA) and the second best.}
\setlength{\tabcolsep}{4.2pt}
\begin{tabular}{lccccccccc}

\toprule
\multirow{2}[3]{*}{Model} & \multicolumn{3}{c}{LLFF}  & \multicolumn{3}{c}{DTU} & \multicolumn{3}{c}{Blender} \\
\cmidrule(lr){2-4}  \cmidrule(lr){5-7} \cmidrule(lr){8-10}

& \(\text{PSNR} \uparrow\) & \(\text{LPIPS} \downarrow\) & \(\text{SSIM} \uparrow\) & \(\text{PSNR} \uparrow\) & \(\text{LPIPS} \downarrow\) & \(\text{SSIM} \uparrow\) & \(\text{PSNR} \uparrow\) & \(\text{LPIPS} \downarrow\) & \(\text{SSIM} \uparrow\) \\

\midrule
Mip-NeRF& 14.62 &0.495 &0.351 &8.68 &0.353 &0.571 & 22.22 & 0.124 & 0.851 \\
DietNeRF& 14.94 &0.496 &0.370 &11.85 &0.314 &0.633 & 23.15  & 0.109& 0.866 \\
RegNeRF&19.08 &0.336 &0.587 &18.89 &0.190 &0.745 &23.83&0.104&0.872 \\
FreeNeRF&19.63 &0.308 &0.612 &{\bf19.92} &0.182 &0.787 &24.26  &0.098&0.883\\
SparseNeRF&{\bf 19.86} &0.328 &\underline{0.624} &\underline{19.55} &0.201 &0.769 &22.41  &0.119&0.861 \\
\midrule
3DGS&16.46 &0.401 &0.440 &14.74 &0.249 &0.672 &22.23 &0.114&0.858\\
DNGaussian&19.12 &\underline{0.294} &0.591 &18.91 &\underline{0.176} &\underline{0.790} &\underline{24.31} &\underline{0.088}&\underline{0.886}\\
UGOT(ours)&\underline{19.77} &{\bf0.273} &{\bf0.625} &19.31 &{\bf0.160} &{\bf0.808} &{\bf24.51}&{\bf0.080}&{\bf0.899} \\

\bottomrule
\end{tabular}
\label{tab:main_result}
\end{table}

\begin{figure}[t]
    \centering
    \includegraphics[width=\linewidth]{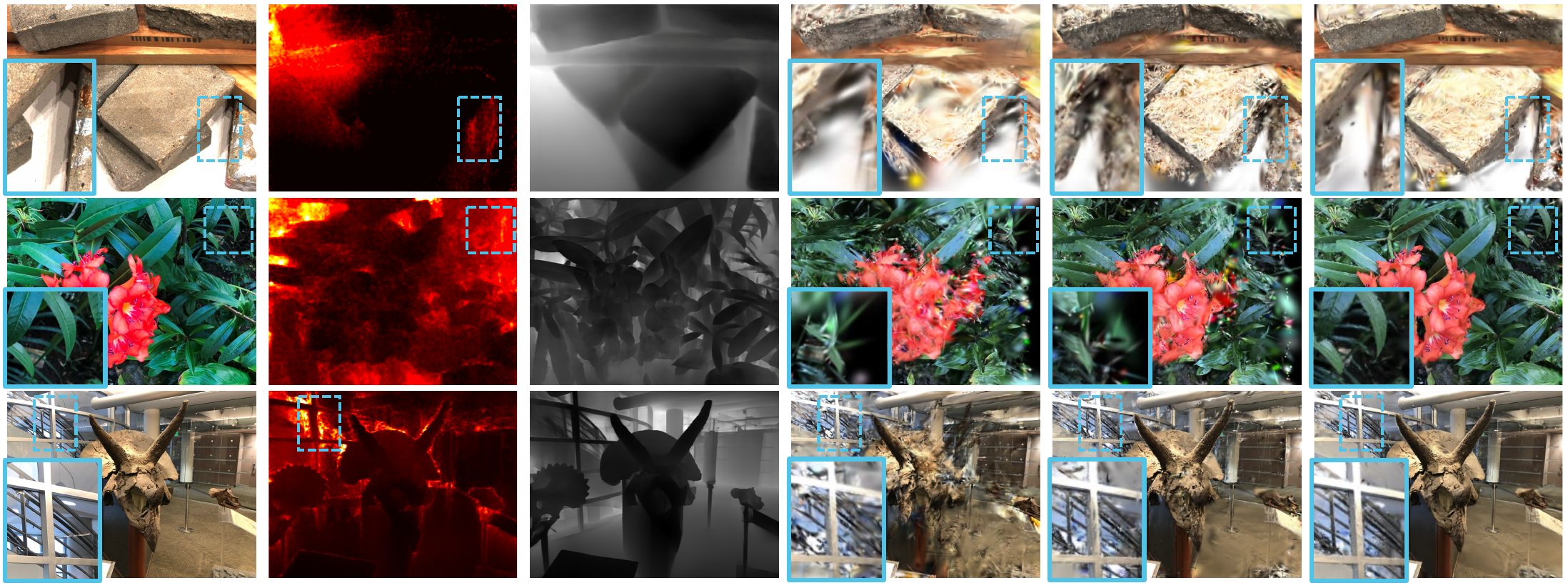}
    \caption{Qualitative results of on DTU and LLFF with 3 input views. From left to right, they are sequentially the original image, the uncertainty map, the depth prediction, 3DGS, DNGaussian and ours.}
    \label{fig:dtu}
\end{figure}

\begin{figure}[t]
    \centering
    \includegraphics[width=\linewidth]{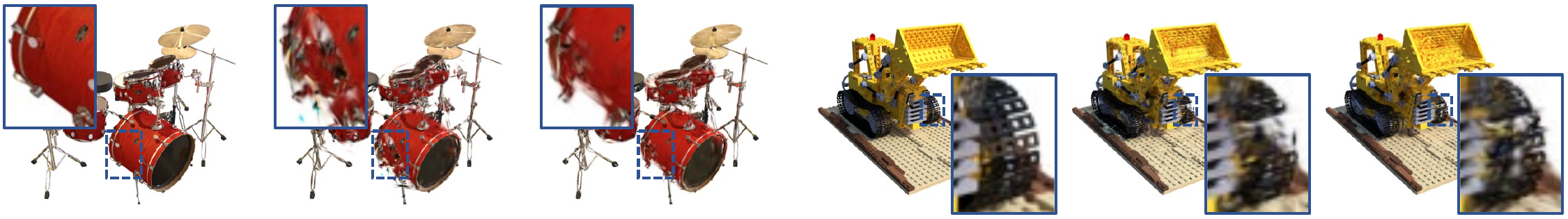}
    \caption{Qualitative results of on Blender with 8 input views. From left to right, they are sequentially the original image, FreeNerf and ours.}
    \label{fig:blender}
\end{figure}

\begin{table}[h!]
  \centering
  \begin{tabular}{@{}l|c|c|c|c|c|c@{}}
    \toprule
    Method & FPS & Time & GPU Mem & PSNR$\uparrow$ & LPIPS$\downarrow$ & SSIM$\uparrow$ \\
    \midrule
             FreeNeRF&  $9 \times 10^{-2}$                 & 2.2 h & $24$ GB         & 19.92 & 0.182 & 0.787 \\
             &                   & 1.1 h & $24$ GB         & 19.80 & 0.189 & 0.781 \\
    \midrule
       SparseNeRF        & $9 \times 10^{-2}$                   & 1.4 h & $12$ GB & 19.55 & 0.201 & 0.769 \\
               &                    & 0.6 h & $12$ GB & 19.45 & 0.207 & 0.763 \\
    \midrule
    DNGaussian & $300$ & $3.6$ min & $2$ GB & 18.91 & 0.176 & 0.790 \\
    \midrule
    Ours & $300$ & $3.8$ min & $2$ GB & 19.31 & 0.160 & 0.808\\
    
    \bottomrule
  \end{tabular}
  \caption{Performance comparison of various methods in terms of frame rate (FPS), processing time, GPU memory usage, and image quality metrics (PSNR, LPIPS, SSIM).}
  \label{tab:speed}
\end{table}

\subsection{Comparison}
\paragraph{DTU Dataset}
From three input views, the qualitative results and visualizations on the DTU dataset are shown in ~\Cref{tab:main_result} and ~\Cref{fig:dtu} first row. Our method achieves the best results in LPIPS and SSIM, and ranked second in PSNR, as displayed in ~\Cref{tab:main_result}. Our scores in PSNR are lower compared to FreeNerf and SparseNerf, primarily due to the inherent weaknesses of 3DGS-based methods in reconstructing untextured backgrounds and voids. Yet, the qualitative examples in ~\Cref{fig:dtu} illustrate that our approach learns more accurate texture information than DNGaussian and restored high-quality details in areas with significant shadows. This is attributed to our method's capability to predict higher uncertainty in shadow regions, utilizing context information to smooth the results instead of pixel-wisely relying on inaccurate depth data.

\paragraph{LLFF Dataset}
In the three-view setting of LLFF, ~\Cref{tab:main_result} shows that our method generally achieves the best results. As the NeRF-based benchmarks interpolate colors into invisible areas, whereas the discrete Gaussian radiance field methods directly expose the black background in these areas, methods based on 3DGS naturally have a disadvantage in reconstructing metrics from these invisible areas. Nevertheless, our approach still surpasses all baselines in the LPIPS and SSIM and are comparable in PSNR to the best methods. ~\Cref{fig:dtu}, rows two and three, presents qualitative outcomes showing our method's ability to predict higher uncertainty in severely defocused (second row) and highly reflective areas along with object edges (third row). Our method employs optimal transport to avoid directly enforcing depth maps to resemble their depth prior, thereby stably enhancing the rendering quality.

\paragraph{Blender Dataset}
We conduct an evaluation on the Blender dataset with eight input views. Our method surpass all other methods in SSIM, LPIPS and even PSNR scores. Although FreeNerf intricately tunes some texture-rich parts through modulation of frequency information, it still fails in areas with complex textures. Our qualitative analysis in ~\Cref{fig:blender} effectively illustrates this point. Our approach leads to less severe artifacts, and crisper and more accurate edges in RGB renderings.

\paragraph{Efficiency}
We further conduct an efficiency study on the DTU 3-view setting with an RTX 3090 GPU to explore the performance of current SOTA baselines. As shown in ~\Cref{tab:speed}, with only 1.1 times the training time, achieves significant performance improvements over DNGaussian. The rendering speed remains unchanged since we are not altering the rendering logic. This also means that, relative to NeRF-based methods, we achieves more than a 1000-fold improvement in rendering speed while nearly matching the performance in quality.

\subsection{Ablation Study}
\begin{table}[t]
  \centering
  \caption{Ablation Study}
  \setlength{\tabcolsep}{5pt}
  \begin{tabular}{@{}l|c|c|c|c|c|c@{}}
    \toprule
    & \multicolumn{3}{c|}{LLFF} & \multicolumn{3}{c}{DTU} \\ 
    \cmidrule(lr){2-4} \cmidrule(lr){5-7} 
    & PSNR $\uparrow$ & LPIPS $\downarrow$ & SSIM $\uparrow$ & PSNR $\uparrow$ & LPIPS $\downarrow$ & SSIM $\uparrow$ \\
    \midrule
    Baseline~\cite{li2024dngaussian} &19.12 &0.294 &0.591 &18.91 &0.176 &0.790 \\
    \hspace{10pt}+pixel-wise OT  & 19.08 & 0.299 & 0.581 & 18.79 & 0.185 & 0.779 \\
    \hspace{10pt}+patch-wise OT  & 19.36 & 0.288 & 0.604 & 19.08 & 0.166 & 0.800 \\
    \hspace{10pt}+uncertainty-guided OT (ours)  &19.77 &0.273 &0.625 &19.31 &0.160 &0.808 \\
    uncertainty-guided OT (w/o L2)  &19.08 &0.306 &0.577 &18.82 &0.190 &0.771   \\
    \bottomrule
  \end{tabular}
  \label{tab:ablation}
\end{table}
\paragraph{Ablation on core components.}
In ~\Cref{tab:ablation}, we set DNGaussian~\cite{li2024dngaussian} (with our depth estimation) as baseline and evaluate the impact of each proposed component on the model's performance. While completely removing the pixel-wise L2 loss does not achieve optimal results, and even performs worse than the baseline, our method effectively mitigates the amplification of local errors in depth prediction that typically occurs in 3D gaussain based methods. By integrating L2 loss, our approach maximizes the utilization of depth information, thus achieving superior performance.

\paragraph{Analysis of TopK values.}
\begin{wraptable}[9]{r}{22em} \small
    \vspace{-3.5mm}  
  \centering
  \caption{TopK values Analysis}
  \begin{tabular}{@{}c|c|c|c|c@{}}  
    \toprule
    TopK& Total weight& PSNR $\uparrow$ & LPIPS $\downarrow$ & SSIM $\uparrow$   \\
    \midrule
    K=1 & 9.8\% &18.72 &309 &0.564  \\
    K=5 & 13.96\%  &18.99 &0.294 &0.580  \\
    K=10 & 27.04\%  &19.27 &0.288 &0.592  \\
    K=20 & 52.91\%  &19.77 &0.273 &0.625  \\
    K=40 & 63.09\%  &19.62 &0.277 &0.619  \\
    \bottomrule
  \end{tabular}
\label{tab:topk}
\end{wraptable}
In ~\Cref{tab:topk}, we further investigate the effectiveness of different TopK values on LLFF. With smaller values of K, back propagation affects fewer Gaussians, and a limited number of Gaussians do not adequately reconstruct the scene's geometry. However, as the value of K increases, the total weight of the most significant Gaussians exceeds 50\%, and updating only these Gaussians during back propagation can already prevent overfitting. Nonetheless, as K continues to increase, the jitter effect of fitting distributions becomes apparent, leading to each pixel's rendering error affecting many Gaussians, which furthur amplifies the error effect in depth prediction.

\paragraph{Analysis of uncertainty.}
\begin{wraptable}[9]{r}{13em} \small
    \vspace{-7.5mm}
    \caption{Uncertainty Analysis}
    \label{tab:uncertainty}
    \begin{tabular}{@{}c|c|c@{}}
        \toprule
        Uncertainty & Sign  & PSNR $\uparrow$ \\
        \midrule
        \multirow{2}{*}{plain} & -  & 18.88  \\
                              & +  & 19.15  \\
        \midrule
        \multirow{2}{*}{exponential} & - &   18.92  \\
                                     & + &  19.49  \\
        \midrule
        \multirow{2}{*}{softmax} & -   &19.22  \\
                                 & +   & 19.77  \\
        \bottomrule
    \end{tabular}
    \label{tab:uncertainty}
\end{wraptable}
In ~\Cref{tab:uncertainty}, we further explore the effectiveness of computed uncertainties on LLFF. Direct use of diffusion-based uncertainty predictions (plain) does not significantly alter performance. However, modulation through exponential or softmax approaches, especially using softmax within patches, yields the best results. By mistakenly using the uncertainty as "certainty" through negation of the signs, we observed a significant drop in performance, further validating the logical soundness of our uncertainty design.

\section{Conclusion}
In this paper, we present the Uncertainty-guided Optimal Transport (UGOT) for depth supervision in sparse-view 3D Gaussian splatting for novel view synthesis. Our approach integrates uncertainty estimates with depth priors to selectively reducing depth supervision in areas of lower uncertainty, and employs an optimal transport framework to align the depth distribution closer to the ground truth. Extensive experiments on the LLFF, DTU, and Blender datasets demonstrate that UGOT significantly outperforms existing methods as well as achieves faster convergence and superior real-time rendering quality. This validates UGOT as an effective solution for enhancing novel view synthesis under conditions of sparse input views and variable depth certainty.

\clearpage 
\bibliographystyle{unsrt}
\bibliography{mybib}

\end{document}